\documentclass[letterpaper, 10 pt, conference]{URL-ieeeconf}
\IEEEoverridecommandlockouts    %
\usepackage{graphics}
\usepackage{times}
\usepackage{amsmath}
\usepackage{amssymb}
\usepackage{graphicx}
\usepackage{algorithm}
\usepackage[noend]{algpseudocode}
\usepackage{booktabs}
\usepackage{color}
\usepackage{multirow}
\usepackage{subcaption}
\usepackage{rotating}
\usepackage{cite}
\definecolor{instructioncolor}{rgb}{.5,.5,.5}

\def\eqref#1{(\ref{#1})}

\newcommand{\rom}[1]{\uppercase\expandafter{\romannumeral #1\relax}}

\makeatletter
\usepackage{xspace}
\DeclareRobustCommand\onedot{\futurelet\@let@token\@onedot}
\def\@onedot{\ifx\@let@token.\else.\null\fi\xspace}

\makeatother

\usepackage{array}
\newcolumntype{L}[1]{>{\raggedright\let\newline\\\arraybackslash\hspace{0pt}}m{#1}}
\newcolumntype{C}[1]{>{\centering\let\newline\\\arraybackslash\hspace{0pt}}m{#1}}
\newcolumntype{R}[1]{>{\raggedleft\let\newline\\\arraybackslash\hspace{0pt}}m{#1}}

\renewcommand{\b}[1]{\mbox{\boldmath$#1$}}

\newcommand{\dR}{\b R}
\newcommand{\dS}{\b S}

\def\dSE3{\mathrm{SE}(3)}
\def\dSO3{\mathrm{SO}(3)}
\def\dS2{\mathbb{S}^2}
\def\dR3{\mathbb{R}^3}
\def\dR6{\mathbb{R}^6}

\usepackage{url}
\usepackage[hidelinks]{hyperref} %

\usepackage{titlesec}
\titlespacing{\section}{0pt}{6pt}{4pt}
\titlespacing{\subsection}{0pt}{4pt}{2pt}

\usepackage{xcolor}

\title{\LARGE \bf SE(3)-LIVOM: Adaptive Multi-LiDAR-Inertial-Visual \\ Odometry and Mapping}

\author{Gunhee Shin$^{1}$, Seungjae Lee$^{1}$, Daebeom Kim$^{1}$, Seoyeon Jang$^{1}$ and Hyun Myung$^{1}$%
  \thanks{$^{1}$School of Electrical Engineering, KAIST (Korea Advanced Institute of Science and Technology), Daejeon, 34141, South Korea, {\tt\scriptsize \{gunhee$\_$shin, sj98lee, ted97k, 9uantum01, hmyung\}@kaist.ac.kr}}
}
\begin{document}
\IEEEaftertitletext{{\centering
    \includegraphics[width=\textwidth]{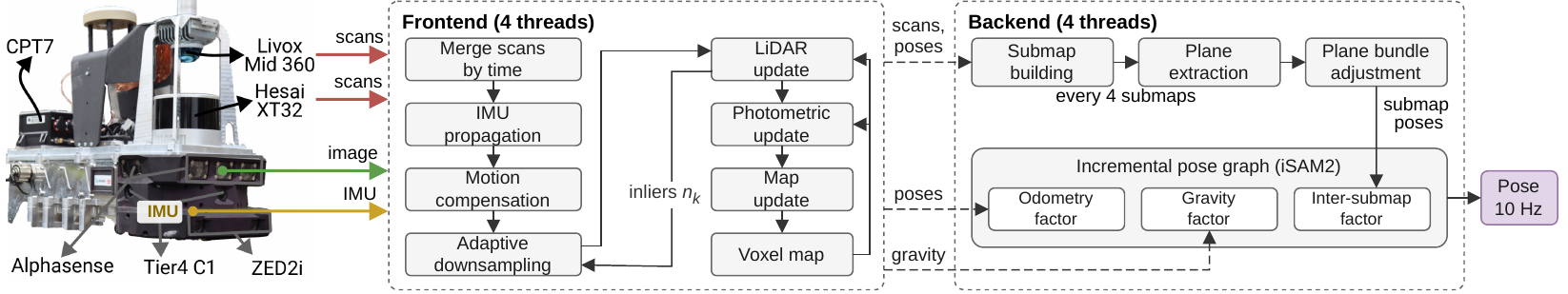}\par
    \captionsetup{font=footnotesize}
    \captionof{figure}{Pipeline of SE(3)\mbox{-}LIVOM. Red, yellow and green lines carry the LiDAR, IMU and camera data, respectively. The frontend and the backend run in separate threads, and dashed arrows pass data between them. The sensor payload image is adapted from\cite{frey2026GrandTour}.}
    \label{fig:framework}
\par}
\vspace{4pt}
}
\maketitle
\thispagestyle{empty}
\pagestyle{empty}

\begin{abstract}
This paper presents SE(3)\mbox{-}LIVOM, the overall winner of the COMFORT Challenge at IROS 2026.
The challenge asks for the trajectory of a legged robot in the GrandTour dataset, whose scenes include collapsed buildings, construction sites and snowfields.
SE(3)\mbox{-}LIVOM runs in real time on two LiDARs, one camera and an IMU.
Its frontend extends SE(3)\mbox{-}LIO to multiple LiDARs and a camera.
Adaptive downsampling keeps the time per scan near 35~ms within and across scenes.
An online backend corrects drift with a plane bundle adjustment over submaps and an incremental pose graph.
With a single configuration for all scenes, the system reaches an average ATE of 1.40~cm on the six test scenes.
\end{abstract}

\section{Introduction}
\label{sec:introduction}

The GrandTour dataset\cite{frey2026GrandTour} records a legged robot in collapsed buildings, on construction sites and on snowfields.
A state estimator for this robot must run in real time in all of these scenes without retuning.
In a narrow corridor, a single LiDAR sees only surfaces along the motion and loses the direction of travel.
On open terrain, a scan holds far more points than inside a building.
A downsampling resolution tuned for one scene is therefore too fine or too coarse for another.
Drift also accumulates over a long mission, because each scan is registered only to the map near it.

FAST-LIVO2\cite{zheng2025FASTLIVO2} fuses one LiDAR, an IMU and a camera in a single filter.
AdaLIO\cite{lim2023AdaLIO} adapts the downsampling resolution to the scene.
X-ICP\cite{tuna2024XICP} detects the directions that a registration cannot constrain.
Hierarchical bundle adjustment\cite{liu2023LargeScale} makes a LiDAR map globally consistent, but it runs offline.
SE(3)\mbox{-}LIVOM builds on our prior work, SE(3)\mbox{-}LIO\cite{shin2026SE3LIO}, and brings these ideas together on a legged robot.
Our contributions are as follows:
\begin{itemize}
    \item We extend SE(3)\mbox{-}LIO to run with multiple LiDARs and a camera.
    \item We propose adaptive downsampling to bound computation time within and across scenes.
    \item We add an online backend with a plane bundle adjustment and an incremental pose graph.
\end{itemize}

SE(3)\mbox{-}LIVOM is the overall winner of the COMFORT Challenge at IROS 2026.
With a single configuration, it reaches an average ATE of 1.40~cm on the six test scenes (Table~\ref{tab:main_results}).
It runs in real time on eight CPU cores, and the code is available at \href{https://github.com/url-kaist/se3-livom-comfort}{\path{https://github.com/url-kaist/se3-livom-comfort}}.

\begin{table*}[t!]
    \captionsetup{font=footnotesize}
    \setlength{\tabcolsep}{3pt}
    \centering
    \caption{ATE RMSE (cm) on the validation and test scenes of the GrandTour dataset. Each validation scene was recorded at the site of the test scene in the same column. Ours (w/o backend) is the frontend alone, and Sub.\ ID is the Codabench submission. \textbf{Bold} is the best and \underline{underline} the second best.}
    {\footnotesize
        \resizebox{\ifdim\width>\textwidth\textwidth\else\width\fi}{!}{%
        \begin{tabular}{l|cccccc|c|cccccc|c|c}
            \toprule
            Split & \multicolumn{7}{c|}{Validation} & \multicolumn{8}{c}{Test} \\
            \cmidrule(lr){2-8} \cmidrule(lr){9-16}
            Scene & $\texttt{ARC-3}$ & $\texttt{ARC-6}$ & $\texttt{CON-3}$ & $\texttt{EIG-2}$ & $\texttt{SNOW-3}$ & $\texttt{SPX-1}$ & Avg. & $\texttt{ARC-2}$ & $\texttt{ARC-7}$ & $\texttt{CON-4}$ & $\texttt{EIG-1}$ & $\texttt{SNOW-2}$ & $\texttt{SPX-2}$ & Avg. & Sub. ID \\
            \midrule
            FAST-LIVO2\cite{zheng2025FASTLIVO2} & 1.34 & 1.24 & 0.97 & 2.59 & 3.44 & 1.64 & 1.87 & 1.22 & 1.38 & 3.13 & 3.32 & 1.61 & 2.28 & 2.16 & 880291 \\
            Ours (w/o backend) & \underline{1.24} & \textbf{0.89} & \underline{1.00} & \underline{2.10} & \underline{1.43} & \underline{1.42} & \underline{1.35} & \underline{0.86} & \underline{1.19} & \underline{2.30} & \underline{2.21} & \textbf{1.41} & \underline{1.99} & \underline{1.66} & 933011 \\
            Ours & \textbf{1.20} & \underline{0.91} & \textbf{0.80} & \textbf{1.70} & \textbf{1.35} & \textbf{1.25} & \textbf{1.20} & \textbf{0.77} & \textbf{1.18} & \textbf{1.68} & \textbf{1.54} & \underline{1.42} & \textbf{1.81} & \textbf{1.40} & 935484 \\
            \bottomrule
        \end{tabular}%
        }
    }
    \vspace{0.1cm}
    \label{tab:main_results}
\end{table*}

\section{Method}
\label{sec:method}
\subsection{Frontend}
\label{subsec:frontend}

\textit{SE(3)\mbox{-}LIO.}
The frontend is built on SE(3)\mbox{-}LIO\cite{shin2026SE3LIO}.
A LiDAR scan is accumulated while the robot moves.
The motion inside the scan is therefore predicted from the IMU first, and the points are then deskewed with that motion.
SE(3)\mbox{-}LIO improves both steps.
For the prediction, it propagates rotation and translation together on the SE(3) manifold, so the predicted pose stays accurate under aggressive motion.
For motion compensation, it treats the predicted motion as uncertain, so the less reliable the pose used to deskew a point, the less that point counts in the update.

\textit{Multiple LiDARs.}
In the corridors of a collapsed building (\mbox{ARC-2} and \mbox{ARC-3}), every visible surface runs along the motion.
With the Hesai LiDAR alone, the frontend loses the direction along the corridor for up to 19~s at a time, as measured with the localizability of X-ICP\cite{tuna2024XICP}.
Adding the downward-looking Livox LiDAR reduces this to at most 3~s.
The Livox also sees the ground near the robot and lowers the error on the metal stairs of \mbox{SPX-1}.
We therefore extend SE(3)\mbox{-}LIO from one LiDAR to two.
The points of all sensors are merged into one scan ordered by time, and each point keeps the noise model of its sensor.
The two LiDARs see the same surfaces at different times, so an error in the predicted motion now appears as a mismatch between the sensors.
The prediction on the SE(3) manifold makes this error smaller, and the point uncertainty reduces the effect of the error that remains.
The LiDARs then act as one sensor with a wider field of view.

\textit{Photometric update.}
After the LiDAR update, the camera image refines the same state with the sparse-direct photometric error of FAST-LIVO2\cite{zheng2025FASTLIVO2}.
Image patches are attached to LiDAR map points, and the update aligns the current image to the patches of the points in view.

\textit{Adaptive downsampling.}
A voxel grid with resolution $l$ downsamples each scan before the update.
A resolution fine enough for narrow buildings leaves too many points on open terrain, where the update slows down.
A resolution coarse enough for open terrain leaves too few matches with the map inside buildings.
AdaLIO\cite{lim2023AdaLIO} adapts the resolution to the scene, and X-ICP\cite{tuna2024XICP} reads localizability from the correspondences of the registration.
We combine the two ideas.
At every scan $k$, the resolution $l_k$ is adjusted so that the number of inliers $n_k$ of the LiDAR update stays near a target $N$ as follows:
\begin{equation} %
    \begin{aligned}
        r_k &= \min\!\left(\max\!\left((n_k/N)^{1/1.5},\, 0.8\right),\, 1.05\right), \\
        l_{k+1} &= \min\!\left(\max\!\left(l_k r_k,\; l_{\min}\right),\; l_{\max}\right)
    \end{aligned}
    \label{eq:adaptive_leaf}
\end{equation}
with $N=2000$, $l_{\min}=0.1$~m and $l_{\max}=0.5$~m.

\begin{figure}[!t]
    \centering
    \includegraphics[width=\columnwidth]{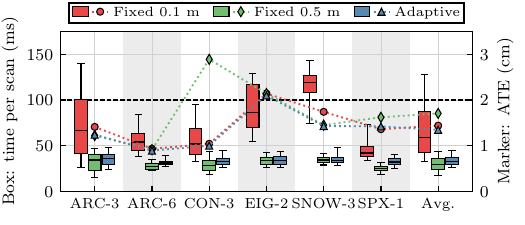}
    \captionsetup{font=footnotesize}
    \caption{Frontend time per scan (boxes, left axis) and ATE (markers, right axis) on the validation scenes with two fixed downsampling resolutions and our adaptive one. Boxes span the interquartile range and the lines the 5th to 95th percentiles. The dashed line is the scan period.}
    \label{fig:adaptive_time}
\end{figure}

\begin{table}[t]
    \captionsetup{font=footnotesize}
    \setlength{\tabcolsep}{3pt}
    \centering
    \caption{Runtime on the test scenes, bags played at real time on 8 cores and 16~GB. A round is one bundle adjustment, and 100~ms is the scan period.}
    {\footnotesize
        \resizebox{\ifdim\width>\columnwidth\columnwidth\else\width\fi}{!}{%
        \begin{tabular}{l|cccccc}
            \toprule
            Scene & $\texttt{ARC-2}$ & $\texttt{ARC-7}$ & $\texttt{CON-4}$ & $\texttt{EIG-1}$ & $\texttt{SNOW-2}$ & $\texttt{SPX-2}$ \\
            \midrule
            Frontend, mean (ms) & 30.1 & 31.8 & 42.3 & 31.6 & 33.5 & 27.8 \\
            Scans over 100~ms & 0 & 0 & 10 & 0 & 0 & 0 \\
            Backend, longest round (s) & 1.4 & 1.4 & 19.4 & 4.4 & 3.1 & 0.7 \\
            Peak memory (GB) & 9.0 & 7.5 & 16.7 & 11.0 & 6.2 & 7.9 \\
            \bottomrule
        \end{tabular}%
        }
    }
    \label{tab:runtime}
\end{table}

\subsection{Backend}
\label{subsec:backend}

\textit{Plane bundle adjustment.}
The frontend registers each scan to the map near it, so the error between poses far apart in time accumulates.
The backend corrects this drift with a global plane bundle adjustment\cite{liu2023LargeScale} over submaps.
Every 25 scans are merged into a submap at their frontend poses and downsampled to 0.2~m.
Whenever four new submaps arrive, the backend runs the bundle adjustment.
The adjustment divides the points of all submaps into 1~m voxels and keeps the voxels whose points form a plane.
It then optimizes the submap poses so that the points in each voxel fit their plane.

\textit{Incremental pose graph.}
The backend output is a pose graph solved with iSAM2\cite{kaess2012ISAM2}.
Frontend poses give odometry factors within a sliding window of 10 scans.
Each bundle adjustment gives an inter-submap factor, the relative pose between every two submaps.
A gravity factor from the mean accelerometer reading is added to the scans where the robot stands still.
The graph is updated incrementally as these factors arrive, and the latest estimate is output as the pose at 10~Hz.

\section{Experiments}
\label{sec:results}

\textit{Dataset.}
We evaluate on the GrandTour dataset\cite{frey2026GrandTour} through the COMFORT benchmark.
The ground truth of the six test scenes is hidden, and the benchmark server scores them.
For each test scene, we also report a validation scene recorded at the same site.
The ground truth is the position of a prism tracked by a Leica MS60 total station.

\textit{Setup.}
We use the Hesai XT32 and Livox Mid-360 LiDARs, the front camera of the Alphasense unit and the STIM320 IMU.
The system runs on eight cores of an Intel Core i9-13900 with 16~GB of RAM.
All scenes share one configuration except the IMU time offset, which is estimated per scene from the gyroscope and the trajectory rotation.
The ATE is the RMSE of the position error in cm after a rigid alignment without scale.

\textit{Accuracy.}
Table~\ref{tab:main_results} lists the ATE on the six test scenes and on their paired validation scenes from the same sites.
The backend lowers the average ATE from 1.66~cm to 1.40~cm on the test scenes.
The improvement is largest on EIG and CON, horizontally on EIG and in height on CON.

\textit{Adaptive downsampling.}
We compare the adaptive resolution with fixed ones in Fig.~\ref{fig:adaptive_time}.
The adaptive resolution matches the accuracy of the best fixed value, while a fixed resolution of 0.5~m triples the error on \mbox{CON-3}.
The adaptive time per scan stays near 35~ms within and across the scenes, whereas a fixed resolution of 0.1~m varies with the scene and exceeds the scan period on most scans of \mbox{SNOW-3} and on many of \mbox{EIG-2}.

\textit{Runtime.}
We assign four threads to the frontend and the other four to the backend.
The frontend takes 28 to 42~ms per scan on the twelve scenes of Table~\ref{tab:main_results}.
Table~\ref{tab:runtime} lists the runtime on the test scenes.
On five of the six test scenes, each bundle adjustment finishes within the 10~s before the next one.
The later rounds on \mbox{CON-4} take up to 19~s, so the correction arrives late but the output is not delayed.

\section{Conclusion}
\label{sec:conclusion}

We presented SE(3)\mbox{-}LIVOM, a multi-LiDAR-inertial-visual odometry and mapping system for legged robots in the field.
Adaptive downsampling keeps the time per scan near 35~ms across scenes.
The online backend lowers the average test ATE from 1.66~cm to 1.40~cm.
On \mbox{CON-4}, a bundle adjustment round takes up to 19~s, so a faster backend for large scenes is left for future work.

\end{document}